\documentclass[hyperref]{article}
\usepackage{amsmath}
\usepackage{graphicx} 
\usepackage{url}
\usepackage{CJKutf8}

\newcommand{\tabincell}[2]{\begin{tabular}{@{}#1@{}}#2\end{tabular}}

\begin{document}
\title{\bf Chinese Lexical Analysis with Deep Bi-GRU-CRF Network} 
\author{\sffamily Zhenyu Jiao, Shuqi Sun$^*$, Ke Sun\\
   {\sffamily\small Natural Language Processing Department, Baidu, Beijing, China}\\
   {\sffamily\small jiaozhenyu@baidu.com }
   {\sffamily\small sunshuqi01@baidu.com }
   {\sffamily\small sunke@baidu.com }
}
\renewcommand{\thefootnote}{\fnsymbol{footnote}}
\footnotetext[1]{Corresponding author.}
\maketitle

{\noindent\small{\bf Abstract:}
 Lexical analysis is believed to be a crucial step towards natural language understanding and has been widely studied.
 Recent years, end-to-end lexical analysis models with recurrent neural networks have gained increasing attention.
 In this report, we introduce a deep Bi-GRU-CRF network that jointly models word segmentation, part-of-speech tagging and named entity recognition tasks. 
 We trained the model using several massive corpus pre-tagged by our best Chinese lexical analysis tool, together with a small, yet high-quality human annotated corpus.
 We conducted balanced sampling between different corpora to guarantee the influence of human annotations, and fine-tune the CRF decoding layer regularly during the training progress.
 As evaluated by linguistic experts, the model achieved a 95.5\% accuracy on the test set, roughly 13\% relative error reduction over our (previously) best Chinese lexical analysis tool. 
 The model is computationally efficient, achieving the speed of 2.3K characters per second with one thread.
}

\begin{CJK*}{UTF8}{gbsn}
\section{Introduction}
In this report, we introduce a Chinese lexical analysis model that jointly accomplishes three tasks: word segmentation, part-of-speech tagging, and named entity recognition.
Most east Asian languages including Chinese are written without explicit word delimiters, therefore, word segmentation is a preliminary step for processing those languages \cite{cai-zhao:2016:P16-1}.  
Part-of-speech (POS) tagging refers to the process of marking each word in the word segmentation result with a correct part of speech, e.g. noun, verb, adjective, etc.
Named entity recognition (NER), refers to recognizing entities that have specific meanings in the identified text, including persons, locations, organization, etc \cite{mohit2014named}.

After over a decade of accumulation and innovation, Baidu have built a series of effective tools for these individual analysis tasks.
Arranges have been made to ensure these tools share the same low-level analysis results (e.g. word segmentations, POS tags) so that developers can integrate multiple tools conveniently \cite{Ritter:2011:NER:2145432.2145595}.
However, such integration works in a pipeline manner that suffers from the error propagation issue, and loses the opportunity to share features between different tasks \cite{qiu-zhao-huang:2013:EMNLP} \cite{wang-zong-xue:2013:Short}.
Besides, after all the endeavors, conflicts still happen when integrating tools with overlapped functionalities.
Sometimes it may cause heavy computation overhead to resolve such conflicts. 

For instance, when building our lexical analysis system published on \emph{Baidu AI open platform}\footnote{\url{https://ai.baidu.com/tech/nlp/lexical}}, we carefully selected models of each sub-task and wrote sophisticated conflict resolution logics to guarantee the quality of the final model.
Eventually, the system outperforms several main-stream competitive products, and receives more than 1 million service invocations per day.
However, because of the simultaneous model calling, the system runs slow, and takes tens of gigabytes of RAM.
Moreover, the complex integration work flows and post-processing logics make the system difficult to optimize.

To tackle with these issues, we re-construct the online system with a single model.
Sequence labeling is a conventional approach to lexical analysis.
The NLP community have paid intensive attention to NN based sequence tagging models recently, and have got encouraging results \cite{zheng-chen-xu:2013:EMNLP} \cite{xu-jiang-watcharawittayakul:2017:Long} \cite{peng-dredze:2016:P16-2}.
The neural structure and its probabilistic output make the model compact and easy to extend or adapt to vertical domains.
We first invoked our online service with 630 million of queries, titles and news sentences, and collect the analysis result as a massive pseudo-annotated corpus.
Afterwards, we trained a deep Bi-GRU-CRF model using both the pseudo-annotated corpus and a small human-annotated corpus.
We oversampled the human-annotated corpus to guarantee its influence to the model.
We also use the label transitions in this corpus to fine-tune the CRF decoding layer during the training progress.

The model works in a full end-to-end manner and turns out to be effective and efficient.
Its input is merely character embedding, without any hand-crafted features.
The model outputs tags according to a unified tag scheme with IOB2-style decoration, thus jointly accomplishes all three analysis tasks.
We invited third-party linguistic experts to perform evaluation on 500 news sentences.
The new model achieves 95.5\% accuracy of both word boundaries and tags, even outperforms the online system to a small extent. 
In terms of efficiency, the new model processes 2.3K characters per second with memory usage less than 100MB.
We have constructed an open-source toolkit based on the model, and have made it public on GitHub\footnote{\url{https://github.com/baidu/lac}}.

\section{Task Specification}
We aim at jointly accomplishing word segmentation, POS tagging, and NER tasks in this report.
For a line of input text, basically a series of characters $c_1, c_2, \cdots, c_T$, our mission is to tag each $c_i$ with a label $l_i$ in the form of ``$t\textrm{-}[\textrm{BI}]$''.
Following the IOB2 format, each tag $t$ is decorated with suffix which is either ``B'' or ``I'', suggesting that a character is the beginning of, or inside a word with tag $t$.
It is worth noting that POS tagging and NER tasks are jointly handled, thus no character would be labeled as ``outside'', i.e. there is no ``O'' suffix.

\begin{table}[!htbp]
	\centering
	\caption{Tag scheme of our lexical analysis result.}
	\label{tagScheme}
	\begin{tabular}{cll}
	\hline
	Tag	&	Description	&	Notes \\
	\hline
	n	&	Noun	&	 \\
	\hline
	PER	&	Person	&	 \\
	\hline
	nr	&	Person	&	Low-confidence person  \\
	\hline
	LOC	&	Location 	&	 \\
	\hline
	ns	&	Location 	&	Low-confidence location  \\
	\hline
	ORG	&	Organization 	&	 \\
	\hline
	nt	&	Organization	&	Low-confidence organization  \\
	\hline
	nw	&	Artwork	&	e.g. 四世同堂 \\
	\hline
	nz	&	Other proper noun	&	 \\
	\hline
	TIME	&	Time	&	 \\
	\hline
	t	&	Time	&	Low-confidence time \\
	\hline
	f	&	Orientation word	&	e.g. 上, 下, 左, 右 \\
	\hline
	s	&	Locative word	&	e.g. 国内, 海外 \\
	\hline
	v	&	Verb	&	 \\
	\hline
	vd	&	Verb used as an adverb	&	 \\
	\hline
	vn	&	Verb used as a noun	&	 \\
	\hline
	a	&	Adjective	&	 \\
	\hline
	ad	&	Adjective used as an adverb	&	 \\
	\hline
	an	&	Adjective used as a noun	&	 \\
	\hline
	d	&	Adverb	&	 \\
	\hline
	m	&	\tabincell{l}{Numeral / \\ Numeral-measure compound}	&	e.g. 一, 第一, 一个 \\
	\hline
	q	&	Measure word	&	 \\
	\hline
	p	&	Preposition	&	 \\
	\hline
	c	&	Conjunction	&	 \\
	\hline
	r	&	Pronoun 	&	 \\
	\hline
	u	&	Auxiliary	&	 \\
	\hline
	xc	&	Other function word	&	\tabincell{l}{Interjection, modal particle, \\ onomatopoeia, and  \\ non-morpheme character} \\
	\hline
	w	&	Punctuation	&	 \\
	\hline
	\end{tabular}
\end{table}

Table \ref{tagScheme} shows the tag scheme we use in the analysis tasks.
For NER task, we aim at recognizing persons (PER), locations (LOC), organizations (ORG), and time (TIME).
For POS tagging task, we develop a compact POS set based on the PKU POS set \cite{yu2000guideline}.
Compared with the latter, we mainly make the following adjustments:

\begin{itemize}
	\item Remove morpheme (*g), front / rear component (h / k). Because we perform a joint analysis, these components are supposed to be part of other words;
	\item Remove temporary phrase (l), idiom (i), and abbreviation (j), and label these words according to their actual function in the sentence;
	\item Remove distinguishing word (b), label it as adjective (a); 
	\item Remove interjection (i), modal particle (y),  onomatopoeia (o), and non-morpheme character (x), label them as ``other function word'' (xc);
	\item Label adjacent numeral and measure word together as a numeral-measure compound (still using the tag ``m'');
	\item Add a new label for artworks (nw).
\end{itemize}

Note that the POS set also contains tags for person, location, organization names and time.
We do not merge them into the corresponding NE tags, and use them to indicate that the tagged NE is with low confidence.
Applications may use this setting to balance between precision and recall of NEs.

\section{Model}
\begin{figure}[!htpb]
	\centering
	\includegraphics[width=0.6\textwidth]{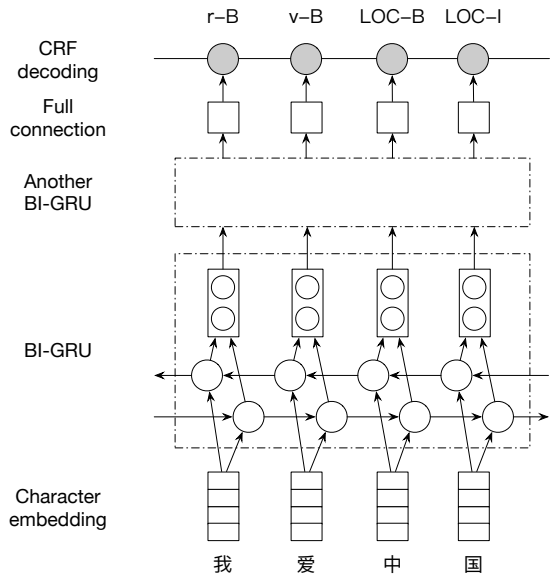}
	\caption{Illustration of our Bi-GRU-CRF network with 2 stacked Bi-GRUs}
	\label{Fig1} 
\end{figure}

In this section, we describe the proposed deep Bi-GRU-CRF model in detail.
Figure \ref{Fig1} illustrates the entire network.

Recurrent Neural Networks (RNN) are important tools for sequence modeling and have been successfully used in several natural language processing tasks. 
Unlike feed-forward neural networks, RNN can model the dependencies between elements of sequences. 
As two main variant of RNN, GRU \cite{chung2014empirical} and LSTM \cite{sak2014long} aim at modeling long-term dependency in long sequences. 
Previous studies have proved that GRU is comparable to LSTM \cite{chung2014empirical}. 
How to choose the gated units in a specific task depends mainly on the data set. 
We found GRU performs better in our experiments. 

Bi-directional GRU (Bi-GRU) is a extension to GRU and is proven suitable for lexical analysis task, in which the input is an entire sentence.
Thus, it is important to have the future encoded as well as the history.
In particular, a reversed direction GRU is combined with a forward GRU to form an Bi-GRU layer. 
These two GRUs take the same input but train in different directions, and concatenate their results as output.
Deep, hierarchical neural networks can be efficient at representing some functions and modeling varying-length dependencies \cite{pascanu2013construct}.
Therefore, stacking multiple Bi-GRUs to form a deep network becomes an inevitable choice for improving the representation capability.
In this report, we stack two Bi-GRUs.

On top of the GRU structures, we use a Conditional Random Field (CRF) \cite{lafferty2001conditional} layer to jointly decode the final label sequence.
Its input is provided by a full connection layer, which transform the topmost Bi-GRU layer's output into a $\mathcal{L}$-dimension vector. $\mathcal{L}$ is the number of all possible labels.
We also try to emphasize the dependencies between output labels \cite{he-EtAl:2017:Long3} by applying hard constraints to the decoding process.
In particular, any sequence that does not comply with IOB2 transfer rules is rejected, such as a ``LOC-B'' followed by a ``TIME-I''.

The model is compact.
It takes less than 100M RAM, including the character embedding lookup table.

\subsection{Stacked Bi-GRUs}
The first step of using neural networks to process symbolic data is to represent them using distributed vectors, also called embeddings \cite{bengio2003neural}.
We only take the original sentence, basically a sequence of characters as input without any hand-crafted features.
Given a character sequence $\{c_1, c_2, \cdots, c_T\}$, each character $c_i$ in the vocabulary $V$ is projected into real-valued vectors $e(c_i)$ through a lookup table.

Afterwards, we build a deep GRU neural network to learn the structural information of a given sentence with the characters embeddings as input.
The GRU is define as follows:
\begin{equation}
u_t = \sigma_{g}(W_{ux}x_{t} + W_{uh}h_{t-1} + b_u) 
\end{equation}
\begin{equation}
r_t = \sigma_g(W_{rx}x_{t} + W_{rh}h_{t-1} + b_r)
\end{equation}
\begin{equation}
\tilde{h_t} = \sigma_{c}(W_{cx}x_{t} + W_{ch}(r_t \odot h_{t-1}) + b_c)
\end{equation}
\begin{equation}
h_t = (1-u_t) \odot h_{t-1} + u_t \odot \tilde{h_t}
\end{equation}
where $\odot$ is the element-wise product of the vector. 
$\sigma_{g}$ is the activation function of update gate $u_t$ and reset gate $r_t$.
$\sigma_{c}$ is the activation function for candidate hidden state. 

Two GRUs which train on the same input sequence in different direction make up a Bi-GRU layer.
Multiple Bi-GRU layers are stacked on top of each other and take the output from lower Bi-GRU layer as the input.

\subsection{CRF with Constrained Decoding}
The CRF layer learns the conditional probability $p(\boldsymbol{y}|\boldsymbol{h})$, where $\boldsymbol{h} = \{h_1, h_2,$ $ \cdots, h_T\}$ are sequences of the representation produced by the topmost Bi-GRU layer and $\boldsymbol{y} = \{y_1, y_2, \cdots, y_T\}$ are label sequences.

The probabilistic model for linear chain CRF defines a family of conditional probability $p(\boldsymbol{y}|\boldsymbol{h};t,s)$ over all possible label sequences $\boldsymbol{y}$ given $\boldsymbol{h}$ with the following form:
\begin{equation}
p(\boldsymbol{y}|\boldsymbol{h};t,s) = \frac{\prod\limits_{i=1}^{T}{\psi_i(y_{i-1}, y_i, \boldsymbol{h})}}{\sum\limits_{\boldsymbol{y}' \in \mathcal{Y}(\boldsymbol{h})} \prod\limits_{i=1}^{T} \psi_i(y_{i-1}', y_{i}', \boldsymbol{h})}
\end{equation}
where $\psi(y_{i-1},y_i,\boldsymbol{h}) = {exp}(\sum_{i=1}^{T}t(y_{i - 1}, y_{i}, \boldsymbol{h}) +s(y_i, \boldsymbol{h}))$ and $ \mathcal{Y}(\boldsymbol{h})$ represents all possible tag sequences (even those that do not comply with the IOB2 format). 
$t$ denotes the transition probabilities from $y_{i-1}$ to $y_{i}$ given input sequence $\boldsymbol{h}$. 
$s$ is the output of the linear function implicitly defined by the full-connection layer, which turns the topmost Bi-RNN's output at time step $i$ to an emission score of $y_i$.

We use maximum conditional likelihood estimation \cite{elkan2008log} to train the CRF layer, with the log-likelihood given by:
\begin{equation}
L(t,s) = \sum_i log\ p(\boldsymbol{y}|\boldsymbol{h}; t,s)
\end{equation}

To decode, simply search via Viterbi algorithm through $\mathcal{Y}(\boldsymbol{h})$ for a sequence that maximizes the conditional probability $P(\boldsymbol{y}|\boldsymbol{h})$, i.e.,
\begin{equation}
\boldsymbol{y}^* = \mathop {\arg \max}_{\boldsymbol{y} \in \mathcal{Y}(\boldsymbol{h})}p(\boldsymbol{y}|\boldsymbol{h})
\end{equation}

As mentioned earlier, we impose constraints during decoding to ensure that the results comply with IOB2 format. 
Label sequence do not satisfy the following conditions are rejected:
\begin{itemize}
	\item The label of the first character of the input sentence cannot be an I-label;
	\item The previous label of each I-label can only be a B-label or a I-label of the same type, e.g., the label before ``LOC-I``can only be ``LOC-B'' or ``LOC-I''.
\end{itemize}
Note that these constraints only affect decoding, so the training speed will not be affected.

\section{Experiments}


In this section, we introduce the details of re-constructing our online lexical analysis system using the proposed Bi-GRU-CRF model.
We first discuss the dataset and detailed training settings, and summarize the comparison results.

\subsection{Dataset}
To get sufficient amount data to let the Bi-GRU-CRF model mimic our online system, we constructed a massive corpus that contains texts form five domains: Web page title, Web search query, news crawled from the Web, and essays from Baidu information flow.
This corpus are pre-labeled using the online system, and works as a pseudo-annotated training set.
To avoid the model over fits to the pseudo labels which contain errors, we introduced a small human-annotated corpus to regularize the training.
Table \ref{differentCorpus} summarizes the details of each corpus in the training set.

\begin{table}[!htbp]
	\centering
	\caption{Composition of the training set}
	\label{differentCorpus}
	\begin{tabular}{lccc}
		\hline Corpus Type & \tabincell{c}{Average\\Length} & Size & Labeled by\\
		\hline
		Web page title & 26 & 264 million & Machine\\
		Web search query & 11& 269 million & Machine\\
		News from Web& 42 & 28 million & Machine\\
		Essay from Baidu information flow & 43 & 68 million & Machine\\
		Human annotated corpus & 23 & 227k & Human\\
		\hline
	\end{tabular}
\end{table}

Different types of the corpora have varied average sentence lengths and distinctive language styles, which helps to improve the robustness of the model.

\subsection{Training}
The network consists of 2 Bi-GRU layers (equal to 2 forward GRUs and 2 reversed GRUs) with 256-dimensional hidden units.
The dimension of character embeddings is set to 128. 
All the weight matrices in embedding, Bi-GRU, full connection layers are initialized with random matrices. 
All elements in the initial matrices are sampled from a uniform distribution with min -0.1 and max 0.1.
For Bi-GRU layer, we use $sigmoid$ for the gate activation function and use $tanh$ for the activation function of candidate hidden state.

Parameter optimization is performed using stochastic gradient descent.
The base learning rate is set to 1e-3.
The learning rate of the embedding layer is set to 5e-3.
The batch size is set to 250, and we select 50 samples from each type of corpus to form each batch of input.
The selection is conducted in a random, non-replacement manner.
If a corpus is exhausted, we restart the selection from the beginning.
This manner ensures the influence of each corpus type, especially for the small human-annotated set.

Due to the huge training dataset, the model is trained for only 1 epoch with early stopping based on the development set.
Specially, after each $10, 000$ batches, we use the entire human-annotated corpus to fine-tune the CRF decoding layer for better transition probabilities.
The whole training stage took about 1 week to finish on single  NVIDIA P40 Card using Baidu's Paddle\footnote{\url{https://github.com/PaddlePaddle/Paddle}} toolkit.

\subsection{Comparisons}
The test set contains of 500 randomly selected sentences of the news crawled from Web during July 2016.
We invited third-party linguistic experts to evaluate the results.
To evaluate the overall quality, we use the accuracy of both boundaries and tags defined as:
\begin{equation}
	Acc = \frac{\#\ of\ correct\ words}{\#\ of\ words\ produced\ by\ the\ system}
\end{equation}
The word is consider ``correct'' if and only if its boundary and tag (including POS and NER tags) are both correct.
In addition, we calculated precision, recall and F1 measures specially for the NER task.

\begin{table}[!htbp]
	\centering
	\caption{The comparison of overall quality.}
	\label{comparisonOnAll}
	\begin{tabular}{cc}
		\hline Model & Acc \\
		\hline
		Online system & 0.948 \\
		Bi-GRU-CRF  & 0.955 \\
		\hline
	\end{tabular}
\end{table}

\begin{table}[!htbp]
	\centering
	\caption{The comparison of NER performance.}
	\label{comparisonOnNER}
	\begin{tabular}{cccc}
		\hline Model & Precision & Recall & F1-Score\\
		\hline
		Online system & 0.916 & 0.829  & 0.871\\
		Bi-GRU-CRF  & 0.903 & 0.854  & 0.878\\
		\hline
	\end{tabular}
\end{table}

Table \ref{comparisonOnAll} and Table \ref{comparisonOnNER} summarize the performance figures of our online system and the Bi-GRU-CRF model.
Results show that on news sentences, a single end-to-end model can achieve comparable or better performance than the online system which combines complex models and heuristics.
Simplified model also benefits the speed.
We built a open-source toolkit based on the Bi-GRU-CRF model using Baidu's Paddle inference API.
The toolkit parses 2.3K characters per second, 57\% faster than the existing online system.

\section{Conclusion}
We introduced a stacked Bi-GRU neural network with CRF decoding layer for a joint lexical analysis task of word segmentation, POS tagging and NER.
We trained the model on a massive dataset pseudo-annotated by the lexical analysis system on \emph{Baidu AI open platform}, together with a human-annotated dataset as regularization.
The resulting model re-constructs the online system well, with a 0.7\% improvement on the overall accuracy, a faster speed, and a tiny memory footprint.
We have built a open-source toolkit based on the model and have made it public on GitHub.
In future work, we will try to further improve the analyzing speed, verify the performance on more domains, and release more models on GitHub.

\end{CJK*}

\bibliographystyle{unsrt} 
\bibliography{reference.bib}

\begin{thebibliography}{10}

\bibitem{cai-zhao:2016:P16-1}
Deng Cai and Hai Zhao.
\newblock Neural word segmentation learning for chinese.
\newblock In {\em Proceedings of the 54th Annual Meeting of the Association for
  Computational Linguistics (Volume 1: Long Papers)}, pages 409--420, Berlin,
  Germany, August 2016. Association for Computational Linguistics.

\bibitem{mohit2014named}
Behrang Mohit.
\newblock Named entity recognition.
\newblock In {\em Natural language processing of semitic languages}, pages
  221--245. Springer, 2014.

\bibitem{Ritter:2011:NER:2145432.2145595}
Alan Ritter, Sam Clark, Mausam, and Oren Etzioni.
\newblock Named entity recognition in tweets: An experimental study.
\newblock In {\em Proceedings of the Conference on Empirical Methods in Natural
  Language Processing}, EMNLP '11, pages 1524--1534, Stroudsburg, PA, USA,
  2011. Association for Computational Linguistics.

\bibitem{qiu-zhao-huang:2013:EMNLP}
Xipeng Qiu, Jiayi Zhao, and Xuanjing Huang.
\newblock Joint {Chinese} word segmentation and {POS} tagging on heterogeneous
  annotated corpora with multiple task learning.
\newblock In {\em Proceedings of the 2013 Conference on Empirical Methods in
  Natural Language Processing}, pages 658--668, Seattle, Washington, USA,
  October 2013. Association for Computational Linguistics.

\bibitem{wang-zong-xue:2013:Short}
Zhiguo Wang, Chengqing Zong, and Nianwen Xue.
\newblock A lattice-based framework for joint chinese word segmentation, pos
  tagging and parsing.
\newblock In {\em Proceedings of the 51st Annual Meeting of the Association for
  Computational Linguistics (Volume 2: Short Papers)}, pages 623--627, Sofia,
  Bulgaria, August 2013. Association for Computational Linguistics.

\bibitem{zheng-chen-xu:2013:EMNLP}
Xiaoqing Zheng, Hanyang Chen, and Tianyu Xu.
\newblock Deep learning for {Chinese} word segmentation and {POS} tagging.
\newblock In {\em Proceedings of the 2013 Conference on Empirical Methods in
  Natural Language Processing}, pages 647--657, Seattle, Washington, USA,
  October 2013. Association for Computational Linguistics.

\bibitem{xu-jiang-watcharawittayakul:2017:Long}
Mingbin Xu, Hui Jiang, and Sedtawut Watcharawittayakul.
\newblock A local detection approach for named entity recognition and mention
  detection.
\newblock In {\em Proceedings of the 55th Annual Meeting of the Association for
  Computational Linguistics (Volume 1: Long Papers)}, pages 1237--1247,
  Vancouver, Canada, July 2017. Association for Computational Linguistics.

\bibitem{peng-dredze:2016:P16-2}
Nanyun Peng and Mark Dredze.
\newblock Improving named entity recognition for chinese social media with word
  segmentation representation learning.
\newblock In {\em Proceedings of the 54th Annual Meeting of the Association for
  Computational Linguistics (Volume 2: Short Papers)}, pages 149--155, Berlin,
  Germany, August 2016. Association for Computational Linguistics.

\bibitem{yu2000guideline}
Shiwen Yu, Xuefeng Zhu, and Huiming Duan.
\newblock The guideline for segmentation and part-of-speech tagging on very
  large scale corpus of contemporary chinese.
\newblock {\em Journal of Chinese Information Processing}, 14(6):58--64, 2000.

\bibitem{chung2014empirical}
Junyoung Chung, Caglar Gulcehre, KyungHyun Cho, and Yoshua Bengio.
\newblock Empirical evaluation of gated recurrent neural networks on sequence
  modeling.
\newblock {\em arXiv preprint arXiv:1412.3555}, 2014.

\bibitem{sak2014long}
Ha{\c{s}}im Sak, Andrew Senior, and Fran{\c{c}}oise Beaufays.
\newblock Long short-term memory recurrent neural network architectures for
  large scale acoustic modeling.
\newblock In {\em Fifteenth annual conference of the international speech
  communication association}, 2014.

\bibitem{pascanu2013construct}
Razvan Pascanu, Caglar Gulcehre, Kyunghyun Cho, and Yoshua Bengio.
\newblock How to construct deep recurrent neural networks.
\newblock {\em arXiv preprint arXiv:1312.6026}, 2013.

\bibitem{lafferty2001conditional}
John~D. Lafferty, Andrew McCallum, and Fernando C.~N. Pereira.
\newblock Conditional random fields: Probabilistic models for segmenting and
  labeling sequence data.
\newblock In {\em Proceedings of the Eighteenth International Conference on
  Machine Learning}, ICML '01, pages 282--289, San Francisco, CA, USA, 2001.
  Morgan Kaufmann Publishers Inc.

\bibitem{he-EtAl:2017:Long3}
Luheng He, Kenton Lee, Mike Lewis, and Luke Zettlemoyer.
\newblock Deep semantic role labeling: What works and what's next.
\newblock In {\em Proceedings of the 55th Annual Meeting of the Association for
  Computational Linguistics (Volume 1: Long Papers)}, pages 473--483,
  Vancouver, Canada, July 2017. Association for Computational Linguistics.

\bibitem{bengio2003neural}
Yoshua Bengio, R{\'e}jean Ducharme, Pascal Vincent, and Christian Janvin.
\newblock A neural probabilistic language model.
\newblock {\em J. Mach. Learn. Res.}, 3:1137--1155, March 2003.

\bibitem{elkan2008log}
Charles Elkan.
\newblock Log-linear models and conditional random fields.
\newblock {\em Tutorial notes at CIKM}, 8:1--12, 2008.

\end{thebibliography}
\end{document}